\documentclass[conference]{IEEEtran}
\IEEEoverridecommandlockouts
\usepackage{cite}
\usepackage{amsmath,amssymb,amsfonts}
\usepackage{algorithmic}
\usepackage{graphicx}
\usepackage{textcomp}
\usepackage{xcolor}
\usepackage[colorlinks,linkcolor=blue]{hyperref}
\def\BibTeX{{\rm B\kern-.05em{\sc i\kern-.025em b}\kern-.08em
    T\kern-.1667em\lower.7ex\hbox{E}\kern-.125emX}}
\begin{document}

\title{Edge-Aware Mirror Network for Camouflaged Object Detection\\
\thanks{*Corresponding author. This work was supported in part by the National Natural Science Foundation of China under the Grant 41927805.}
}

\author{\IEEEauthorblockN{1\textsuperscript{st} Dongyue Sun}
\IEEEauthorblockA{\textit{College of Computer} \\
\textit{Science and Technology}\\
\textit{Ocean University of China}\\
Qingdao, China \\
ssdy@stu.ouc.edu.cn}
\and 
\IEEEauthorblockN{2\textsuperscript{nd} Shiyao Jiang}
\IEEEauthorblockA{\textit{College of Computer} \\
\textit{Science and Technology} \\
\textit{Ocean University of China}\\
Qingdao, China \\
jsy@stu.ouc.edu.cn}
\and
\IEEEauthorblockN{3\textsuperscript{rd} Lin Qi*}
\IEEEauthorblockA{\textit{College of Computer } \\
\textit{Science and Technology} \\
\textit{Ocean University of China}\\
Qingdao, China  \\
qilin@ouc.edu.cn}
}

\maketitle

\begin{abstract}
Existing edge-aware camouflaged object detection (COD) methods normally output the edge prediction in the early stage. However, edges are important and fundamental factors in the following segmentation task. Due to the high visual similarity between camouflaged targets and the surroundings, edge prior predicted in early stage usually introduces erroneous foreground-background and contaminates features for segmentation. To tackle this problem, we propose a novel Edge-aware Mirror Network (EAMNet), which models edge detection and camouflaged object segmentation as a cross refinement process. More specifically, EAMNet has a two-branch architecture, where a segmentation-induced edge aggregation module and an edge-induced integrity aggregation module are designed to cross-guide the segmentation branch and edge detection branch. A guided-residual channel attention module which leverages the residual connection and gated convolution finally better extracts structural details from low-level features. Quantitative and qualitative experiment results show that EAMNet outperforms existing cutting-edge baselines on three widely used COD datasets. Codes are available at \href{https://github.com/sdy1999/EAMNet}{https://github.com/sdy1999/EAMNet}.
\end{abstract}


\section{Introduction}
Camouflage refers to the phenomenon that wild animals adapt their colors and textures to surroundings in order to hide themselves and deceive other animals (\textit{e.g.}, predators). Camouflaged object detection (COD)~\cite{b1} aims to search and segment camouflaged objects from single image. Analyzing and exploring camouflage patterns can benefit a range of downstream applications such as polyp segmentation~\cite{b2} and recreational art~\cite{b3}. Camouflages exhibit highly similarity with background in visual appearance, making COD more challenging than ordinary object detection task, where traditional hand-crafted algorithms can hardly deal with~\cite{b4}.

With the release of large-scale and well annotated COD datasets~\cite{b1,b5,b15}, many deep learning based COD models have been proposed. Fan~\textit{et al.}~\cite{b1} constructed the COD10K dataset which contains 5066 samples of camouflaged objects and proposed a search-identification network (SINet). SINet first uses a search module to roughly locate the camouflaged object, and then uses an identification module for precise segment. Inspired by the design ethos of SINet, a variety of approaches focusing on cross-level feature fusion have been proposed~\cite{b6,b7}. By leveraging the well-designed fusion approaches, these methods can roughly locate regions containing the camouflaged object, but still fail to clarify the indistinct boundary between the camouflaged object and its surroundings, resulting in fuzzy segmentations. 

\begin{figure}[t]
\begin{minipage}[b]{.31\linewidth}
  \centering
\centerline{\includegraphics[width=0.8\textwidth,height=1\textwidth,clip]{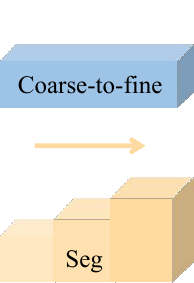}}
  \centerline{(a)}\medskip
\end{minipage}
\hfill
\begin{minipage}[b]{0.31\linewidth}
  \centering
\centerline{\includegraphics[width=0.8\textwidth,height=1\textwidth,clip]{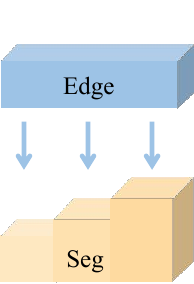}}
  \centerline{(b)}\medskip
\end{minipage}
\hfill
\begin{minipage}[b]{0.31\linewidth}
  \centering
\centerline{\includegraphics[width=0.8\textwidth,height=1\textwidth,clip]{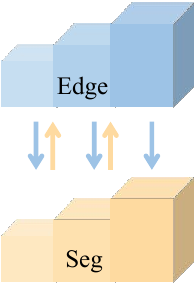}}
  \centerline{(c)}\medskip
\end{minipage}

\begin{minipage}[b]{1\linewidth}
  \centering
\centerline{\includegraphics[width=1\textwidth,height=0.2\textwidth,clip]{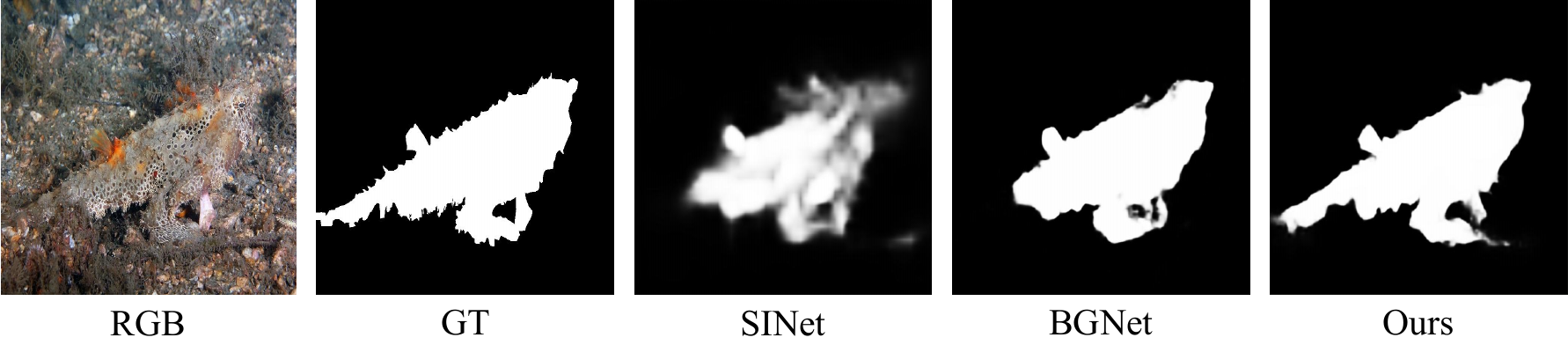}}
  \centerline{(d)}\medskip
\end{minipage}

\caption{Model paradigm (a), represented by SINet~\cite{b1}, focuses on exploring multi-level segmentation feature fusion; model paradigm (b), represented by BGNet~\cite{b13}, introduces edge cues to further enhance representation of segmentation features; model paradigm (c), represented by this approach, models edge detection and camouflaged object segmentation as a cross-refinement process. Our model outperforms the first two models in the integrity of the camouflaged object segmentation (d).}
\label{fig:1}
\end{figure}

To overcome the above limitation, researchers have introduced edge cues to enhance the representation of segmentation features~\cite{b12,b13}. As shown in Figure~\ref{fig:1}(b), the common thread is to generate an edge prediction at early stage of the network and use it as edge prior to guide the fusion of multi-level segmentation features. However, these methods still suffer from two shortcomings: (1) low-level features that preserve low-frequency texture information are less focused. (2) overemphasis on injecting the edge prior into segmentation features without considering the accuracy of early generated edge prediction. Due to the lack of semantics information, edge prediction generated early is prone to lose the integrity of the camouflaged object, which misleads segmentation into erroneous foreground prediction. Following ICON~\cite{b8}, we employ the integrity concept as the whole body of a certain camouflaged object. As shown in Figure~\ref{fig:1}(d), compared to the results of SINet, the prediction of BGNet has clearer boundaries, but still fails to identify the whole body of the batfish.

 In this paper, we propose EAMNet, which consists of an edge detection branch and a segmentation branch. The design ethos of our EAMNet is to utilize edge features to enhance the segmentation features, and then by leveraging the enhanced segmentation features we can explore a more complete foreground representation, which in turn can be utilized to enhance edge features. Two components are proposed to implement the cross guidance at multiple scales: (1) the edge-induced integrity aggregation (EIA) module introduces edge stream to enhance segmentation features and explores the integrity of camouflaged objects from channel and spatial dimensions. (2) the segmentation-induced edge aggregation (SEA) module introduces segmentation stream to help the edge branch for the whole shape of the camouflaged object. Moreover, with the assumption that correct semantic guidance and low-frequency preserved residual connection are the two essential factors for leveraging low-level features, we propose the guided-residual channel attention (GCA) module to extract the structure details of camouflaged objects preserved in low-level features. Experimental results show that the proposed EAMNet surpasses recent nine state-of-the-art methods under four widely used metrics on three COD datasets.

\begin{figure*}[t]
\centerline{\includegraphics[width = \textwidth, height=0.5\textwidth,  clip ]{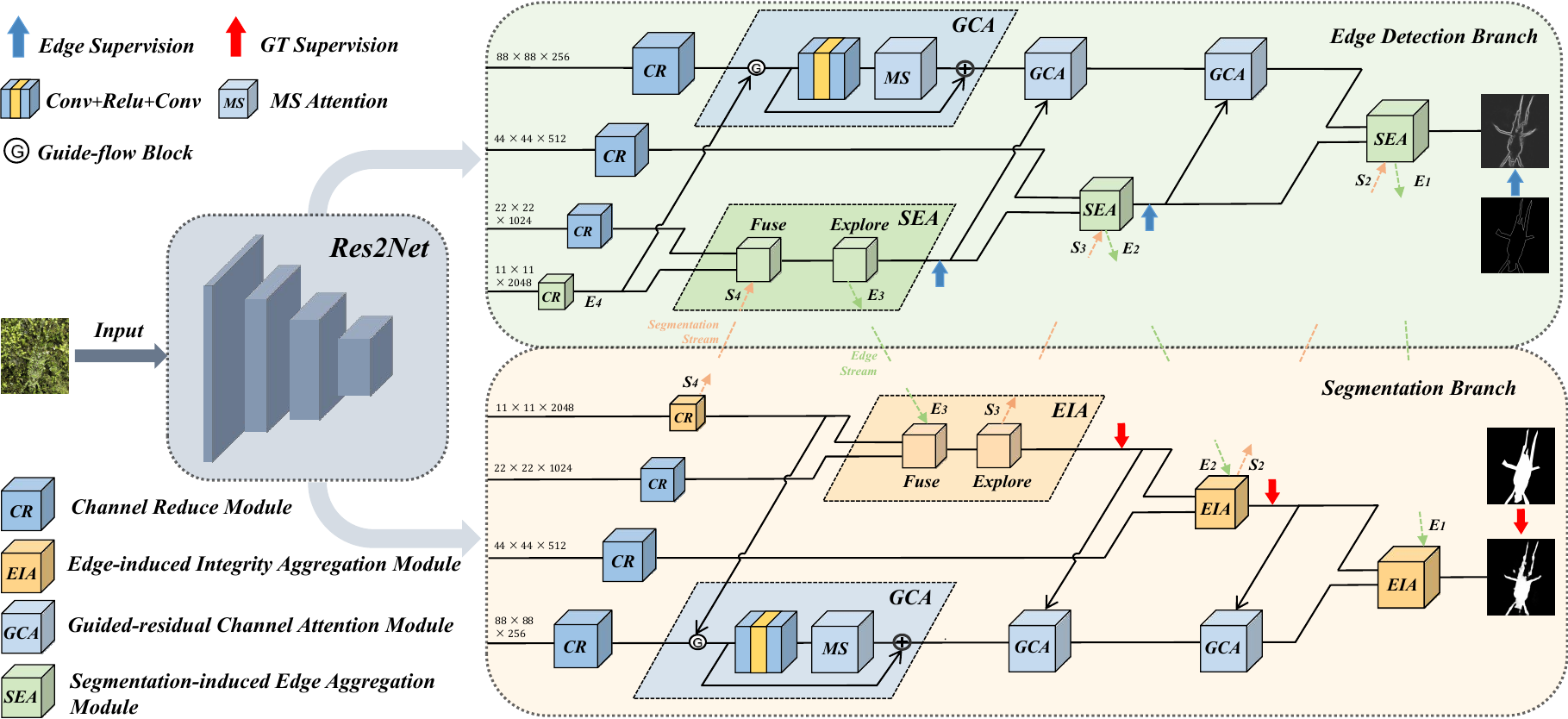}}
\caption{The overall architecture of EAMNet, which featured by three key components, \textit{i.e.}, segmentation-induced edge aggregation module (SEA), edge-induced integrity aggregation module (EIA) and guided-residual channel attention module (GCA).}
\label{fig2}
\end{figure*}

\section{RELATED WORKS}

For COD, traditional methods mostly separate the camouflaged object and its surroundings by utilizing hand-crafted features such as texture, color, and brightness, and these methods often failed to deal with complex scenes~\cite{b4}. 

Recently, deep learning-based approaches have become dominant in COD field~\cite{b1,b6,b9}.
Fan~\textit{et al.}~(SINet)~\cite{b1} designs a search module and an identification module to detect camouflaged objects by simulating the process of animal hunting. Based on the observation that people usually search for camouflaged objects by changing the viewpoint of the same scene, Yan~\textit{el at.}~(MirrorNet)~\cite{b14} introduces a flipped image stream to better locate potential camouflaged objects. Except for these bio-inspired methods, Chen~\textit{et al.}~(C$^2$Net)~\cite{b6} designs an attention-induced cross-level fusion module to fully capture valuable context information to boost the accuracy of COD.
Yang~\textit{et al.}~(UGTR)~\cite{b30} integrates the benefits of Bayesian learning and transformer reasoning to leverage the uncertainty and probabilistic information in COD data annotation. To further clarify the indistinct boundaries,  Sun~\textit{et al.}~\cite{b13} designs an edge-guidance feature module to embed the edge cues into segmentation features. Zhu~\textit{et al.}~\cite{b12} utilizes adaptive space normalization to do this in a more effect way. Notably, there are also some salient object detection (SOD) methods~\cite{b18,b19} that explores the cross guidance between edge detection and salient object segmentation, but their design ethos are detecting the salient object which have strong distinctions to its surroundings, and due to the essential difference between ``camouflaged" and ``salient", it is difficult for applying them into COD field.

\section{METHODOLOGY}

\subsection{Overview}

Figure~\ref{fig2} shows the overview of the proposed EAMNet, which consists of three kinds of key components including the bifurcated backbone, the edge detection branch, and the segmentation branch. The edge detection branch consists mainly of the SEA module, which aggregates multi-level edge features under the guidance from segmentation branch. The segmentation branch consists mainly of the EIA module, which aggregates multi-level segmentation features under the the guidance from edge detection branch. Both the two branches adopt the coarse-to-fine strategy under the supervision of edge map and ground truth map, respectively, which means that the prediction in two branches is progressively optimized from low to high resolution. 

Specifically, given an input image I, we first adopt the Res2Net-50~\cite{b20} as backbone to extract features at four levels, which can be denoted as $ F = \{f_{i}, i = 1,2,3,4\}$. Then, we fed F into channel reduce modules containing a $1\times1$ convolutional layer to extract multi-level edge features with the channel size of 64 denoted as $F_e=\{f^e_{i},i=1,2,3,4\}$ and multi-level segmentation features denoted as $F_s=\{f^s_{i},i=1,2,3,4\}$. 
After that, We stack three SEA modules in the edge detection branch to aggregate the multi-level edge features $F_e$ in a coarse-to-fine manner and three EIA modules in the segmentation branch to aggregate segmentation features $F_s$.
For clearer descriptions, We define the output of SEA modules as $E=\{E_{i},i=1,2,3\}$, $f^e_{4}$ as $E_{4}$, the output of EIA modules as $S=\{S_{i},i=1,2,3\}$ and $f^s_{4}$ as $S_{4}$. Multiple SEA modules and EIA modules are interacted in a cascaded manner to implement the cross refinement between the two branches. Moreover, in both two branches, the shallow detail features $f^s_{1}$ and $f^e_{1}$ are firstly fed into three cascaded GCA modules to filter the abundant background noise by leveraging the semantic stream from corresponding aggregation modules. We take the last EIA module's prediction as the final segmentation result for testing stage.

\subsection{Edge-Induced Integrity Aggregation Module}

\begin{figure}[t]
\centerline{\includegraphics[scale = 0.85,clip]{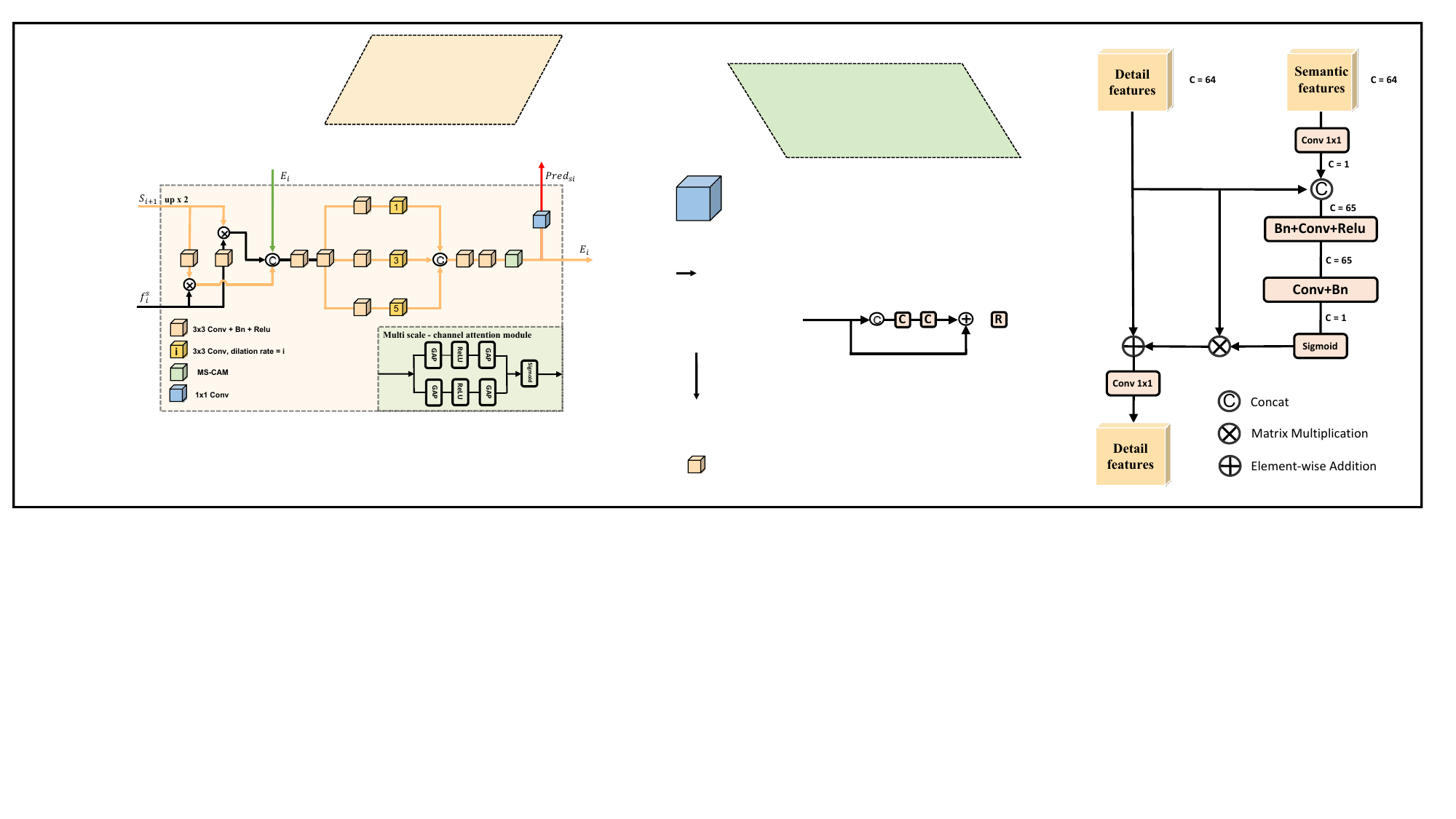}}
\caption{Illustration of the EIA module.}
\label{fig3_EIA}
\end{figure}

The EIA module aims to inject the edge cues into the representation learning of segmentation features, and further explore the integrity information from both channel and spatial dimensions. Figure~\ref{fig3_EIA} shows the detail structure of EIA module, which consists of two stages, the first for feature fusion and the second for the integrity exploration.

Specifically, for the \textit{ith} EIA module, given the segmentation feature $f^s_i\in{R^{{W_{1}} \times {H_{1}} \times 64}}$, the higher level segmentation feature $S_{i+1}\in{R^{{W_{2}} \times {H_{2}} \times 64}}$ from the previous EIA module and the edge feature $E_i\in{R^{{W_{1}} \times {H_{1}} \times 64}}$ from the previous SEA module as inputs.  Inspired by~\cite{b23}, we first employ a mirror multiplication strategy consisting of two paths to fully aggregate the two-level segmentation features $f^s_i$ and $S_{i+1}$ (after up-sampling). In the main path, the higher level feature $S_{i+1}$ is fed into a $3\times3$ convolutional layer to generate a semantic mask. This mask is then multiplied with $f^s_i$ to enhance the response of the camouflaged object. The mirror path, on the other hand, utilizes the lower level feature $f^s_i$ to generate a detail mask. This detail mask is then multiplied with $S_{i+1}$ to preserve fine-grained details that might have been lost in the coarser segmentation map. The two paths are concatenated with the edge feature $E_i$ and fed into a fusion block containing two $3\times3$ convolutional layers to implement the guidance from the edge branch to the segmentation branch. Note that, the $3\times3$ convolutional layer in this paper consists of a $3\times3$ convolution, a batch normalization and a relu function. We denote the the final fused feature as $\widetilde{f^s_i}$, and the whole fusion process can be formulated as:

\begin{equation}
\widetilde{f^s_i} = C_{3\times3}(\textit{Concat}(\mathcal{M}(f^s_i,S_{i+1}),E_i)),
\label{eq:EIA_1}
\end{equation}

{\noindent}After that, the fused feature $\widetilde{f_{s}}$ is fed into a three-branch structure $\textit{B} = \{B_j, j = 1,2,3\}$ to mine the integrity cues from a multi-scale perspective. Each branch contains a $3\times3$ convolutional layer and a $3\times3$ atrous convolutional layer with a dilation rate of $n_j$. In this paper, we set $n_j = \{1,3,5\}$. The three branches are concatenated and fed into two cascaded $3\times3$ convolutional layers for reducing the channels to 64. To further enhance the response of critical channels which preserve weak integrity cues of the camouflaged object (\textit{e.g.}, the legs of a crab), we adopt a multi-scale channel attention module ($\textit{MS}$)~\cite{b29}, which has a two-branch architecture. The attention matrix $M$ of $\textit{MS}$ can be calculated as: $M(X) = L(X) + G(X)$, G(X) aims to discover global information by leveraging the global average pooling (GAP) operation, while L(X) adopt the point-wise convolution to obtain local contexts. and the whole exploration process can be formulated as:

\begin{equation}
{S_i} = \textit{MS}(C_{3\times3}(\textit{Concat}(B_1(\widetilde{f^s_i}),B_2(\widetilde{f^s_i}),B_3(\widetilde{f^s_i})))),
\label{eq:EIA_2}
\end{equation}

{\noindent}where $S_i$ denotes the final output feature of the \textit{ith} EIA module, which will be separately fed into the next SEA module to provide the edge branch with integrity cues, the next EIA module to implement the coarse-to-fine fusion and the corresponding GCA module for background noise filtering. Noted that, with a $1\times1$ convolution to change the channels of feature $S_i$, we obtain the segmentation prediction $P^s=\{P^s_i,i=1,2,3\}$ of the camouflaged object.

\subsection{Segmentation-Induced Edge Aggregation Module}
The SEA module aims to inject the integrity cues into the representation learning of edge features, which shares the same design ethos as the EIA module, but is much lighter. 

Specifically, for the \textit{ith} SEA module, given the edge feature $f^e_i\in{R^{{W_{1}} \times {H_{1}} \times 64}}$, the higher level edge feature $E_{i+1}\in{R^{{W_{2}} \times {H_{2}} \times 64}}$ from the previous SEA module, the segmentation feature $S_{i+1}\in{R^{{W_{2}} \times {H_{2}} \times 64}}$ from the EIA module as inputs. The two-level edge features $f^e_i$ and $E_{i+1}$ (after up-sampling) are concatenated and fed into a fusion block containing two $3\times3$ convolutional layers for fully fusion, and then we adopt the mirror multiplication strategy $\mathcal{M}$ followed by an additional residual connection to inject the integrity cues from segmentation feature $S_{i+1}$ (after up-sampling) into the fused edge feature $f^e_i$. We denote the the final fused feature as $\widetilde{f^e_i}$, the whole fusion process can be formulated as: 

\begin{equation}
\begin{split}
\begin{aligned}
\left\{\begin{array}{lll}
f^e_i &=& C_{3\times3}(Concat(f^e_i,E_{i+1})) , \\
\widetilde{f^e_i} &=& \mathcal{M}(f^e_i, S_{i+1}) + f^e_i.
\end{array}\right.
\end{aligned}
\end{split}
\label{eq:SIE_fuse}
\end{equation}

{\noindent}After that, we feed $\widetilde{f^e_i}$ into two cascaded $3\times3$ convolutional layers to explore its discriminative representation in foreground-background blurred regions, and the final output feature $E_i$ is fed into a $1\times1$ convolution for channel reducing to obtain the edge prediction $P^e=\{P^e_{i},i=1,2,3\}$ of the camouflaged object.

\subsection{Guided-Residual Channel Attention Module}

In natural images, camouflaged objects tend to show smaller dimensions, which makes the structure details preserved in low-level features essential for detecting the integrity of camouflaged objects. Therefore, we propose the Guided-Residual Channel Attention (GCA) module and its structure is illustrated in Figure~\ref{fig2}. The GCA module first applies a guide-flow block that guides semantic features from SEAs (EIAs in segmentation branch) to help filtering the background noise in the detail feature, then uses a channel attention block to further capture the inter dependencies between channels of the detail feature.

\begin{figure}[t]
\centerline{\includegraphics[scale = 0.9]{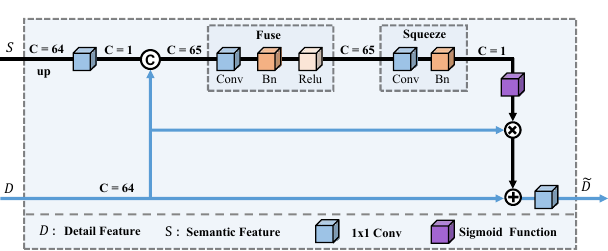}}
\caption{Illustration of the Guide-flow block.}
\label{fig4_guideflow}
\end{figure}

As shown in Figure~\ref{fig4_guideflow}, the guide-flow block takes detail features $D\in{R^{W_1\times H_1\times 64}}$ and semantic features $S\in{R^{W_2{\times}H_2{\times}64}}$ as inputs. The guidance map $G\in{R^{W_1\times H_1\times 1}}$ is obtained as: 
\begin{equation}
G = \textit{Sigmoid}(C_{1\times1}(Concat(D,C_{1\times1}(S)))),
\label{eq:guideflow}
\end{equation}
where $C_{1\times1}$ denotes the normalized $1\times1$ convolutional layers. The guidance maps generated by both $S$ and $D$ can fully utilize the positional information of camouflaged objects in the semantic feature $S$ , while also taking into account the structural information of the camouflaged objects preserved in the detail feature $D$. The new detail feature  $\widetilde{D}\in{R^{W_1\times H_1\times C}}$ is calculated as:
\begin{equation}
\widetilde{D} = C_{1\times1}((G \cdot D) + D),
\label{eq:GRC}
\end{equation}
where $\cdot$ denotes the element-wise product. $C_{1\times1}$ denotes the $1\times1$ convolutional layer, then we fed it into a convolutional block followed by a multi-scale channel attention module $\textit{}{MS}$ to further explore its foreground representation in the channel dimension. Moreover, an additional residual connection is utilized to preserve the low-frequency structure information of camouflaged objects to the maximum extent possible. The process can be formulated as:
\begin{equation}
D =  \textit{MS}(C_{3\times3}( \widetilde{D} ) )+ \widetilde{D},
\label{eq:GCA_ResConnection}
\end{equation}

{\noindent}where $C_{3\times3}$ denotes two $3\times3$ convolutions with a relu function in the middle, $D$ denotes the final output detail feature.

\subsection{Loss Function}


We utilise the co-supervision strategy to jointly train the two branchs. Following the previous state-of-the-art COD methods~\cite{b1,b13}, we adopt the weighted binary cross-entropy loss ($L^w_{BCE}$) and the weighted intersection-over-union loss ($L^w_{IOU}$) for the supervision of camouflaged mask ($M_c$), the dice loss ($L_{dice}$) for the supervision of edge mask ($M_e$). The total loss of EAMNet can be calculated as: $L_{total} = \sum_{i=1}^{3} {\lambda}_i(L^w_{BCE}(M_c,P^s_i) + L^w_{IOU}(M_c,P^s_i) + L_{dice}(M_e,P^e_i))$, where ${\lambda}_i$ is a trade-off parameter and we set ${\lambda}_i = \frac{1}{2^{i-1}}$, $P^s_i$ is the segmentation prediction and $P^e_i$ is the edge prediction of camouflaged object.

\begin{table*}[]
\caption{Quantitative comparison with state-of-the-art methods for COD on three benchmarks using four evaluation metrics (\textit{i.e.}, $S_{\alpha}$, $E_{\phi}$, $F^w_{\beta}$ and $M$), "$\uparrow$" / "$\downarrow$" indicates that higher/lower is better. Top two results are highlighted in \textcolor{red}{red} and \textcolor{blue}{blue}.}
\label{tab:compare}
\resizebox{\linewidth}{!}{
\begin{tabular}{|l|l|llll|llll|llll|}
\hline
\multicolumn{1}{|c|}{} &
  \multicolumn{1}{|c|}{} &
  \multicolumn{4}{c|}{CAMO} &
  \multicolumn{4}{c|}{COD10K} &
  \multicolumn{4}{c|}{NC4K} \\ \cline{3-14}

\multicolumn{1}{|c|}{Method} &
  \multicolumn{1}{|c|}{Pub./Year}  &
  $S_{\alpha}\uparrow$ &
  $E_{\phi}\uparrow$ &
  $F^w_{\beta}\uparrow$ &
  $M\downarrow$ &
  $S_{\alpha}\uparrow$ &
  $E_{\phi}\uparrow$ &
  $F^w_{\beta}\uparrow$ &
  $M\downarrow$ &
  $S_{\alpha}\uparrow$ &
  $E_{\phi}\uparrow$  &
  $F^w_{\beta}\uparrow$ &
  $M\downarrow$  \\ \hline

SINet\cite{b1} & 
  CVPR'20 &
  0.745 &
  0.804 &
  0.644 &
  0.092 &
  0.776 &
  0.864 &
  0.631 &
  0.043 &
  0.808 &
  0.871 &
  0.723 &
  0.058 \\

S-MGL\cite{b17} &
  CVPR'21 &
  0.772 &
  0.806 &
  0.664 &
  0.089 &
  0.811 &
  0.844 &
  0.654 &
  0.037 &
  0.829 &
  0.862 &
  0.731 &
  0.055\\
R-MGL\cite{b17} &
  CVPR'21 &
  0.775 &
  0.812 &
  0.673 &
  0.088 &
  0.814 &
  0.851 &
  0.666 &
  0.035 &
  0.833 &
  0.867 &
  0.739 &
  0.053\\
UGTR\cite{b30} &
  ICCV'21 &
  0.784 &
  0.821 &
  0.683 &
  0.086 &
  0.817 &
  0.852 &
  0.665 &
  0.036 &
  0.839 &
  0.874 &
  0.746 &
  0.052\\
SINet-V2\cite{b9} &
  TPAMI'21  &
  \textbf{\textcolor{blue}{0.820}} &
  \textbf{\textcolor{blue}{0.882}} &
  0.743 &
  \textbf{\textcolor{blue}{0.070}} &
  0.815 &
  0.887 &
  0.680 &
  0.037 &
  0.847 &
  0.903 &
  0.770 &
  0.048 \\
DCTNet\cite{b7} &
  TMM'22 &
  0.778 &
  0.804 &
  0.667 &
  0.084 &
  0.790 &
  0.821 &
  0.616  &
  0.041 &
  - &
  - &
  - &
  - \\
$C_2$-Net\cite{b6} &
  TCSVT'22  &
  0.800 &
  0.869  &
  0.730 &
  0.077 &
  0.811 &
  0.891 &
  0.691 &
  0.036 &
  - &
  - &
  - &
  -\\
BSANet\cite{b12} &
  AAAI'22  &
  0.796 &
  0.851 &
  0.717 &
  0.079 &
  0.818 &
  0.891 &
  0.699 &
  0.034 &
  - &
  - &
  - &
  - \\ 
BGNet\cite{b13} &
  IJCAI'22  &
  0.812 &
  0.870 &
  \textbf{\textcolor{blue}{0.749}} &
  0.073 &
  \textbf{\textcolor{blue}{0.831}} &
  \textbf{\textcolor{blue}{0.901}} &
  \textbf{\textcolor{blue}{0.722}} &
  \textbf{\textcolor{blue}{0.033}} &
  \textbf{\textcolor{blue}{0.851}} &
  \textbf{\textcolor{blue}{0.907}} &
  \textbf{\textcolor{blue}{0.788}} &
  \textbf{\textcolor{blue}{0.044}}\\ \hline
Ours &
    &
  \textbf{\textcolor{red}{0.831}} &
  \textbf{\textcolor{red}{0.890}} &
  \textbf{\textcolor{red}{0.763}} &
  \textbf{\textcolor{red}{0.064}} &
  \textbf{\textcolor{red}{0.839}} &
  \textbf{\textcolor{red}{0.907}} &
  \textbf{\textcolor{red}{0.733}} &
  \textbf{\textcolor{red}{0.029}} &
  \textbf{\textcolor{red}{0.862}} &
  \textbf{\textcolor{red}{0.916}} &
  \textbf{\textcolor{red}{0.801}} &
  \textbf{\textcolor{red}{0.040}} \\ \hline
\end{tabular}
}
\end{table*}

\begin{table*}[]
\caption{Ablation study on GCA module and cross refinement process, '$w/o$' denotes 'without', EDB denotes the whole edge detection branch. The best results are highlighted in \textbf{bold}.
}
\label{table2}
\resizebox{\linewidth}{!}{
\begin{tabular}{|l|llll|llll|llll|}
\hline
\multicolumn{1}{|c|}{} &
  \multicolumn{4}{c|}{CAMO} &
  \multicolumn{4}{c|}{COD10K} &
  \multicolumn{4}{c|}{NC4K} \\ \cline{2-13}

\multicolumn{1}{|c|}{Model} &

  $S_{\alpha}\uparrow$ &
  $E_{\phi}\uparrow$ &
  $F^w_{\beta}\uparrow$ &
  $M\downarrow$ &
  $S_{\alpha}\uparrow$ &
  $E_{\phi}\uparrow$ &
  $F^w_{\beta}\uparrow$ &
  $M\downarrow$ &
  $S_{\alpha}\uparrow$ &
  $E_{\phi}\uparrow$  &
  $F^w_{\beta}\uparrow$ &
  $M\downarrow$  \\ \hline

a. $w/o$ eGCA& 
  0.827 &
  0.890 &
  0.764 &
  0.065 &
  
  0.836 &
  0.911 &
  0.733 &
  0.030 &
   
  0.856 &
  0.913 &
  0.793 &
  0.043 \\ 
  
b. $w/o$ sGCA& 
  0.827 &
  0.886 &
  0.766 &
  0.067 &
  
  0.835 &
  0.908 &
  0.731 &
  0.030 &
  
  0.857 &
  0.915 &
  0.797 &
  0.043 \\

c. $w/o$ eGCA \& sGCA& 
  0.823 &
  0.883 &
  0.757 &
  0.067 &
  0.832 &
  0.906 &
  0.723 &
  0.030 &
  0.852 &
  0.910 &
  0.785 &
  0.045 \\ 
  
d. $w/o$ EDB &
  0.820 &
  0.887 &
  0.758 &
  0.069 &
  
  0.825 &
  0.894 &
  0.708 &
  0.032 &
  
  0.848 &
  0.905 &
  0.778 &
  0.046\\ \hline
e. Ours &

  \textbf{0.831} &
  \textbf{0.890} &
  \textbf{0.763} &
  \textbf{0.064} &
  \textbf{0.839} &
  \textbf{0.907} &
  \textbf{0.733} &
  \textbf{0.029} &
  \textbf{0.862} &
  \textbf{0.916} &
  \textbf{0.801} &
  \textbf{0.040} \\ \hline
\end{tabular}
}
\end{table*}

\section{EXPERIMENTS}

\subsection{Experimental Setup}

\noindent\textbf{Datasets.}  We employ EAMNet on three benchmark datasets: COD10K~\cite{b1}, NC4K~\cite{b5}, CAMO~\cite{b15}. COD10K is the most challenging dataset by far, which consists of 78 subclasses with 5,066 samples. CAMO is also a widely used COD dataset consisting of 1,250 samples. Following the data partition of SINet~\cite{b1}, we use 3,040 samples from COD10K and 1,000 samples from CAMO for training stage, and rest ones for testing stage. Also, we use the NC4K dataset (4,121 samples) to evaluate the generalization ability of EAMNet.

\noindent\textbf{Implementation Details.}  The proposed EAMNet is implemented with pytorch. Following the training settings in recent methods~\cite{b1,b6,b9}, Res2Net~\cite{b20} pretrained on ImageNet is employed as the backbone. The input images are resized to $384 \times 384$ and augmented by randomly horizontal flipping. AdamW with weight decay is chosen as the optimizer. The learning rate is set to 5e-5 and follows a linear warm-up and linear decay strategy which divided by 10 every 50 cycles. The entire model is trained for 100 epochs with a batch size of 24 on a single NVIDIA 3090 GPU.

\noindent\textbf{Evaluation Metrics.}  We apply four evaluation metrics  widely used in COD task, including S-measure ({$S_{\alpha}$})\cite{b26}, E-measure ({$E_{\phi}$})\cite{b27}, weighted F-measure ($F^w_{\beta}$)\cite{b28} and Mean Absolute Error ({$M$})\cite{b31}. In general, a better COD method will present higher {$S_{\alpha}$}, {$E_{\phi}$}, {$F^w_{\beta}$} and lower {$M$}.

\subsection{Performance Comparison}

Table~\ref{tab:compare} shows the quantitative comparison between EAMNet and 9 state-of-the-art methods in terms of $S_{\alpha}$, $E_{\phi}$, $F^w_{\beta}$ and $M$, and it can be observed that our EAMNet performs better than these methods in terms of all metrics. Specifically, in CAMO, our EAMNet outperforms the two edge-aware methods BGNet~\cite{b13} and BSANet~\cite{b12} by 3.5\% and 1.9\% in terms of structural similarity measure $S_{\alpha}$, respectively. This suggests that our EAMNet excels in mining the complete structure of the camouflaged object. Our improved performance can be attributed to the unique two-branch architecture, which allows for more accurate edge prior mining, and the EIA module, which can fully aggregate edge prior and segmentation features in both spatial and channel dimensions. Additionally, the proposed GCA module can make full use of the low-level features to boost COD performance.

\begin{figure*}[t]
\centerline{\includegraphics[width = \textwidth, clip ]{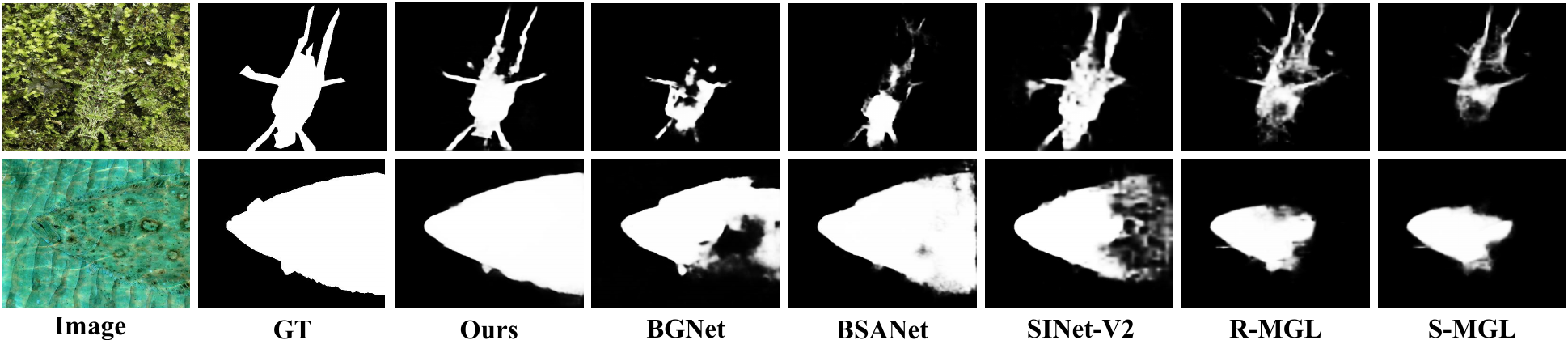}}
\caption{Visual comparison of the proposed EAMNet with five state-of-the-art COD methods in tough scenarios.}
\label{fig_comparison}
\end{figure*}

We further evaluated the generalization ability of our EAMNet by testing it on the NC4K dataset, where it outperformed the second best method, BGNet, by 1.1\%, 0.9\%, and 1.3\% in terms of $S_{\alpha}$, $E_{\phi}$, and $F^w_{\beta}$, respectively. Figure~\ref{fig_comparison} provides visual comparisons between our EAMNet and the other methods in challenging scenes. Our method demonstrates superior performance in accurately detecting and segmenting the whole shape of the camouflaged object, as shown in the example of the Katydid image in the first row. Our EAMNet achieves the structural similarity measure $S_{\alpha}$ of 85\%, while BGNet achieves only 60\%.

\subsection{Ablation study}

To verify the effectiveness of the overall cross refinement strategy, we replace the whole edge detection branch with the boundary detection module in BSANet~\cite{b12}, which generates edge features only in the early stage using the multi-level features extracted by Res2Net. As shown in Table~\ref{table2}, compared to model d, model e achieves obvious performance improvements on all benchmark datasets, with the performance gains of $1.3\%$, $0.9\%$, $1.7\%$ on average in terms of $S_{\alpha}$, $E_{\phi}$ and $F^w_{\beta}$. This is due to the lack of integrity information in the edge prior generated only early, and the cooperation between EIA module and SEA module in our EAMNet is beneficial to solve this problem. In addition, to verify the effectiveness of our GCA module, we remove the three GCA modules in the segmentation branch (sGCA) and the edge detection branch (eGCA), respectively. It can be observed that compared to module e, module c shows a significant decrease in all metrics, which indicates that the proposed GCA module's powerful extraction of structural features in low-level features.

\section{CONCLUSION}

In this paper, we propose a novel framework based on the cross refinement between edge detection and camouflaged object segmentation, called EAMNet. In particular, the EIA and SEA module are proposed to implement the cross-refinement relations. Moreover, we design the GCA module to better extracts structural details from low-level features to boost the accuracy of camouflaged object detection. Extensive experiments on three benchmark datasets have shown that our EAMNet outperforms other state-of-the-art COD methods. In the future, we plan to further explore our EAMNet on weakly supervised COD datasets.

\bibliographystyle{IEEEbib}
\bibliography{EAMNet}

\begin{thebibliography}{10}

\bibitem{b1}
Deng-Ping Fan, Ge-Peng Ji, Guolei Sun, Ming-Ming Cheng, Jianbing Shen, and Ling
  Shao,
\newblock ``Camouflaged object detection,''
\newblock in {\em Proceedings of the IEEE/CVF conference on computer vision and
  pattern recognition}, 2020, pp. 2777--2787.

\bibitem{b2}
Deng-Ping Fan, Ge-Peng Ji, Tao Zhou, Geng Chen, Huazhu Fu, Jianbing Shen, and
  Ling Shao,
\newblock ``Pranet: Parallel reverse attention network for polyp
  segmentation,''
\newblock in {\em Medical Image Computing and Computer Assisted
  Intervention--MICCAI 2020: 23rd International Conference, Lima, Peru, October
  4--8, 2020, Proceedings, Part VI 23}. Springer, 2020, pp. 263--273.

\bibitem{b3}
Ranran Feng and Balakrishnan Prabhakaran,
\newblock ``Facilitating fashion camouflage art,''
\newblock in {\em Proceedings of the 21st ACM international conference on
  Multimedia}, 2013, pp. 793--802.

\bibitem{b4}
Ajoy Mondal,
\newblock ``Camouflaged object detection and tracking: A survey,''
\newblock {\em International Journal of Image and Graphics}, vol. 20, no. 04,
  pp. 2050028, 2020.

\bibitem{b5}
Yunqiu Lv, Jing Zhang, Yuchao Dai, Aixuan Li, Bowen Liu, Nick Barnes, and
  Deng-Ping Fan,
\newblock ``Simultaneously localize, segment and rank the camouflaged
  objects,''
\newblock in {\em Proceedings of the IEEE/CVF Conference on Computer Vision and
  Pattern Recognition}, 2021, pp. 11591--11601.

\bibitem{b15}
Trung-Nghia Le, Tam~V Nguyen, Zhongliang Nie, Minh-Triet Tran, and Akihiro
  Sugimoto,
\newblock ``Anabranch network for camouflaged object segmentation,''
\newblock {\em Computer vision and image understanding}, vol. 184, pp. 45--56,
  2019.

\bibitem{b6}
Geng Chen, Si-Jie Liu, Yu-Jia Sun, Ge-Peng Ji, Ya-Feng Wu, and Tao Zhou,
\newblock ``Camouflaged object detection via context-aware cross-level
  fusion,''
\newblock {\em IEEE Transactions on Circuits and Systems for Video Technology},
  vol. 32, no. 10, pp. 6981--6993, 2022.

\bibitem{b7}
Wei Zhai, Yang Cao, HaiYong Xie, and Zheng-Jun Zha,
\newblock ``Deep texton-coherence network for camouflaged object detection,''
\newblock {\em IEEE Transactions on Multimedia}, 2022.

\bibitem{b13}
Yujia Sun, Shuo Wang, Chenglizhao Chen, and Tian-Zhu Xiang,
\newblock ``Boundary-guided camouflaged object detection,''
\newblock {\em arXiv preprint arXiv:2207.00794}, 2022.

\bibitem{b12}
Hongwei Zhu, Peng Li, Haoran Xie, Xuefeng Yan, Dong Liang, Dapeng Chen,
  Mingqiang Wei, and Jing Qin,
\newblock ``I can find you! boundary-guided separated attention network for
  camouflaged object detection,''
\newblock in {\em Proceedings of the AAAI Conference on Artificial
  Intelligence}, 2022, vol.~36, pp. 3608--3616.

\bibitem{b8}
Mingchen Zhuge, Deng-Ping Fan, Nian Liu, Dingwen Zhang, Dong Xu, and Ling Shao,
\newblock ``Salient object detection via integrity learning,''
\newblock {\em IEEE Transactions on Pattern Analysis and Machine Intelligence},
  2022.

\bibitem{b9}
Deng-Ping Fan, Ge-Peng Ji, Ming-Ming Cheng, and Ling Shao,
\newblock ``Concealed object detection,''
\newblock {\em IEEE Transactions on Pattern Analysis and Machine Intelligence},
  vol. 44, no. 10, pp. 6024--6042, 2021.

\bibitem{b14}
Jinnan Yan, Trung-Nghia Le, Khanh-Duy Nguyen, Minh-Triet Tran, Thanh-Toan Do,
  and Tam~V Nguyen,
\newblock ``Mirrornet: Bio-inspired camouflaged object segmentation,''
\newblock {\em IEEE Access}, vol. 9, pp. 43290--43300, 2021.

\bibitem{b30}
Fan Yang, Qiang Zhai, Xin Li, Rui Huang, Ao~Luo, Hong Cheng, and Deng-Ping Fan,
\newblock ``Uncertainty-guided transformer reasoning for camouflaged object
  detection,''
\newblock in {\em Proceedings of the IEEE/CVF International Conference on
  Computer Vision}, 2021, pp. 4146--4155.

\bibitem{b18}
Zhe Wu, Li~Su, and Qingming Huang,
\newblock ``Stacked cross refinement network for edge-aware salient object
  detection,''
\newblock in {\em Proceedings of the IEEE/CVF international conference on
  computer vision}, 2019, pp. 7264--7273.

\bibitem{b19}
Runmin Wu, Mengyang Feng, Wenlong Guan, Dong Wang, Huchuan Lu, and Errui Ding,
\newblock ``A mutual learning method for salient object detection with
  intertwined multi-supervision,''
\newblock in {\em Proceedings of the IEEE/CVF conference on computer vision and
  pattern recognition}, 2019, pp. 8150--8159.

\bibitem{b20}
Shang-Hua Gao, Ming-Ming Cheng, Kai Zhao, Xin-Yu Zhang, Ming-Hsuan Yang, and
  Philip Torr,
\newblock ``Res2net: A new multi-scale backbone architecture,''
\newblock {\em IEEE transactions on pattern analysis and machine intelligence},
  vol. 43, no. 2, pp. 652--662, 2019.

\bibitem{b23}
Zuyao Chen, Qianqian Xu, Runmin Cong, and Qingming Huang,
\newblock ``Global context-aware progressive aggregation network for salient
  object detection,''
\newblock in {\em Proceedings of the AAAI conference on artificial
  intelligence}, 2020, vol.~34, pp. 10599--10606.

\bibitem{b29}
Yimian Dai, Fabian Gieseke, Stefan Oehmcke, Yiquan Wu, and Kobus Barnard,
\newblock ``Attentional feature fusion,''
\newblock in {\em Proceedings of the IEEE/CVF Winter Conference on Applications
  of Computer Vision}, 2021, pp. 3560--3569.

\bibitem{b17}
Aixuan Li, Jing Zhang, Yunqiu Lv, Bowen Liu, Tong Zhang, and Yuchao Dai,
\newblock ``Uncertainty-aware joint salient object and camouflaged object
  detection,''
\newblock in {\em Proceedings of the IEEE/CVF Conference on Computer Vision and
  Pattern Recognition}, 2021, pp. 10071--10081.

\bibitem{b26}
Deng-Ping Fan, Ming-Ming Cheng, Yun Liu, Tao Li, and Ali Borji,
\newblock ``Structure-measure: A new way to evaluate foreground maps,''
\newblock in {\em Proceedings of the IEEE international conference on computer
  vision}, 2017, pp. 4548--4557.

\bibitem{b27}
Deng-Ping Fan, Cheng Gong, Yang Cao, Bo~Ren, Ming-Ming Cheng, and Ali Borji,
\newblock ``Enhanced-alignment measure for binary foreground map evaluation,''
\newblock {\em arXiv preprint arXiv:1805.10421}, 2018.

\bibitem{b28}
Ran Margolin, Lihi Zelnik-Manor, and Ayellet Tal,
\newblock ``How to evaluate foreground maps?,''
\newblock in {\em Proceedings of the IEEE conference on computer vision and
  pattern recognition}, 2014, pp. 248--255.

\bibitem{b31}
Federico Perazzi, Philipp Kr{\"a}henb{\"u}hl, Yael Pritch, and Alexander
  Hornung,
\newblock ``Saliency filters: Contrast based filtering for salient region
  detection,''
\newblock in {\em 2012 IEEE conference on computer vision and pattern
  recognition}. IEEE, 2012, pp. 733--740.

\end{thebibliography}

\end{document}